\documentclass[sigconf]{acmart}

\newcommand{\VisualAE}{\textbf{VisualAE }}
\newcommand{\NumAE}{\textbf{NumAE }}
\newcommand{\RandomWalk}{\textbf{RandomWalk }}
\newcommand{\ARIMA}{\textbf{ARIMA }}

\usepackage{multirow}
\usepackage{lipsum}
\AtBeginDocument{%
  \providecommand\BibTeX{{%
    \normalfont B\kern-0.5em{\scshape i\kern-0.25em b}\kern-0.8em\TeX}}}

\copyrightyear{2021}
\acmYear{2021}
\setcopyright{acmcopyright}\acmConference[ICAIF'21]{2nd ACM International Conference on AI in Finance}{November 3--5, 2021}{Virtual Event, USA}
\acmBooktitle{2nd ACM International Conference on AI in Finance (ICAIF'21), November 3--5, 2021, Virtual Event, USA}
\acmPrice{15.00}
\acmDOI{10.1145/3490354.3494387}
\acmISBN{978-1-4503-9148-1/21/11}

\begin{document}
\title{Visual Time Series Forecasting: An Image-driven Approach}
\author{Srijan Sood}
\authornote{Both authors contributed equally to this research.}
\affiliation{%
  \institution{J.~P.~Morgan AI Research}
  \city{New York}
  \state{NY}
  \country{USA}}
\email{srijan.sood@jpmorgan.com}

\author{Zhen Zeng}
\authornotemark[1]
\affiliation{%
  \institution{J.~P.~Morgan AI Research}
  \city{New York}
  \state{NY}
  \country{USA}}
\email{zhen.zeng@jpmorgan.com}

\author{Naftali Cohen}
\affiliation{%
  \institution{J.~P.~Morgan AI Research}
  \city{New York}
  \state{NY}
  \country{USA}}
\email{naftali.cohen@jpmorgan.com}

\author{Tucker Balch}
\affiliation{%
  \institution{J.~P.~Morgan AI Research}
  \city{New York}
  \state{NY}
  \country{USA}}
\email{tucker.balch@jpmorgan.com}

\author{Manuela Veloso}
\affiliation{%
  \institution{J.~P.~Morgan AI Research}
  \city{New York}
  \state{NY}
  \country{USA}}
\email{manuela.veloso@jpmorgan.com}
\renewcommand{\shortauthors}{Sood and Zeng, et al.}





\begin{abstract}
Time series forecasting is essential for agents to make decisions. Traditional approaches rely on statistical methods to forecast given past numeric values. In practice, end-users often rely on visualizations such as charts and plots to reason about their forecasts. Inspired by practitioners, we re-imagine the topic by creating a novel framework to produce visual forecasts, similar to the way humans intuitively do. In this work, we leverage advances in deep learning to extend the field of time series forecasting to a visual setting. We capture input data as an image and train a model to produce the subsequent image. This approach results in predicting distributions as opposed to pointwise values. We examine various synthetic and real datasets with diverse degrees of complexity. Our experiments show that visual forecasting is effective for cyclic data but somewhat less for irregular data such as stock price. Importantly, when using image-based evaluation metrics, we find the proposed visual forecasting method to outperform various numerical baselines, including ARIMA and a numerical variation of our method. 
We demonstrate the benefits of incorporating vision-based approaches in forecasting tasks – both for the quality of the forecasts produced, as well as the metrics that can be used to evaluate them.

\end{abstract}

\maketitle

\section{Introduction and Related Work}\label{sec:related}
Time series forecasting is a standard statistical task that concerns predicting future values given historical information.
Conventional forecasting tasks range from uncovering simple periodic patterns to forecasting intricate nonlinear patterns. 
The prevailing and most widely used forecasting techniques include linear regression, exponential smoothing, and ARIMA (e.g., \cite{friedman2001elements,hyndman2018forecasting,makridakis2020m4}). 
In recent years, modern approaches emerge as tree-based algorithms, ensemble methods, neural network autoregression, and recurrent neural networks (e.g., \cite{friedman2001elements}). These methods are useful for highly nonlinear and inseparable data but are often considered less stable than the more traditional approaches (e.g., \cite{hyndman2018forecasting,krispin2019hands}). 

In the last few years, deep learning approaches have been applied in the domain of time series analysis, for forecasting~\cite{romeu2015stacked, gensler2016deep, bao2017deep, sagheer2019time}, as well as unsupervised approaches for pre-training, clustering, and distance calculation~\cite{abid2018learning, mousavi2019unsupervised, sagheer2019unsupervised, tavakoli2020clustering}. The common theme across these works is their use of stacked autoencoders (with different variations -- vanilla, convolutional, recurrent, etc.) on numeric time series data.
Autoencoders have also shown promise in the computer vision domain across tasks as image denoising~\cite{alain2014regularized, gondara2016medical}, image compression~\cite{Akyazi_2019_CVPR_Workshops}, and image completion and in-painting~\cite{mao2016image, li2017generative}.

\begin{figure}[tb!]
    \centering
    \includegraphics[width=0.385\textwidth]{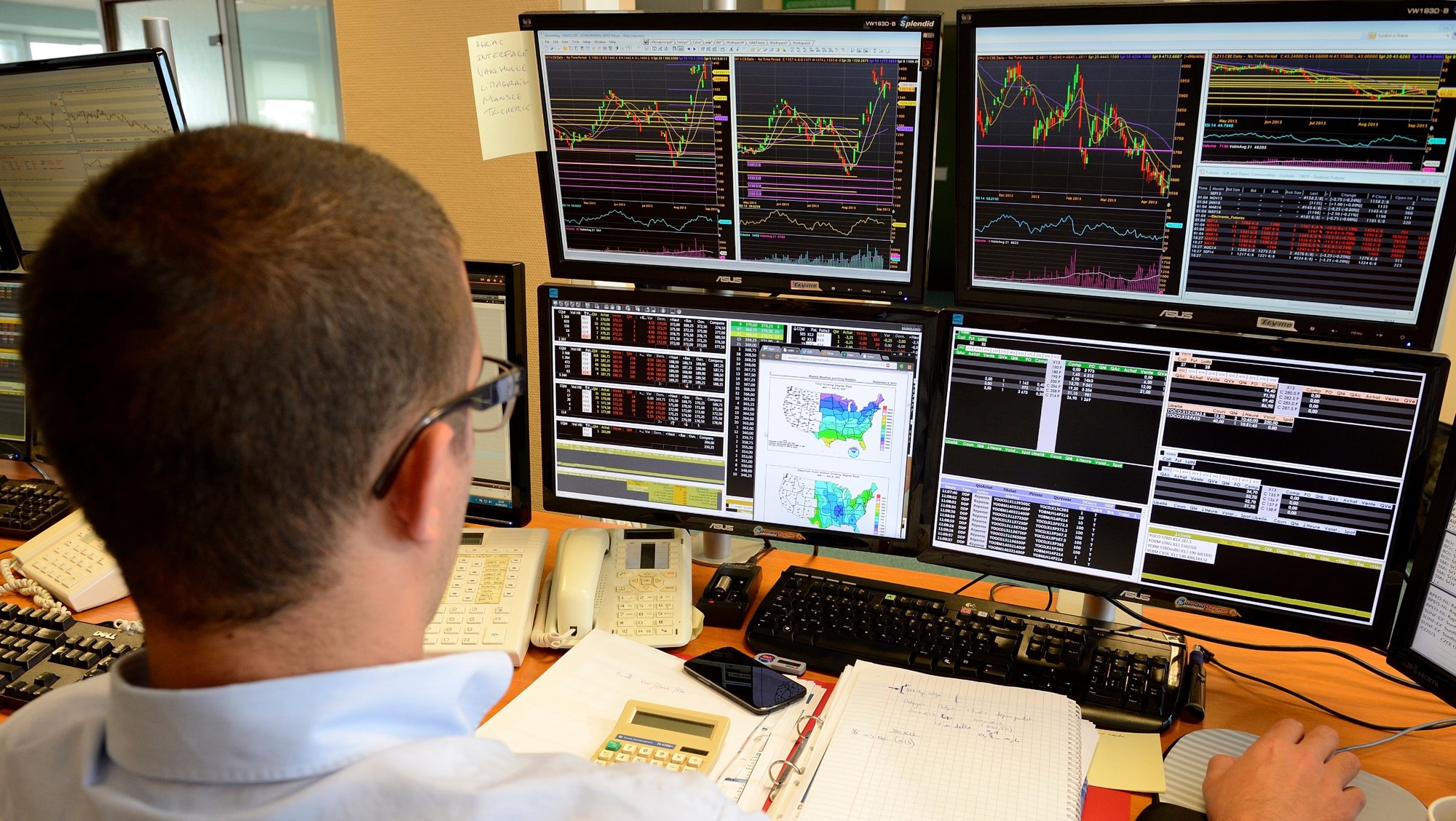}
    \caption{Typical workstation of a professional trader. Credit: Photoagriculture/ Shutterstock.com}
    \label{fig:computer}
    \vspace{-6mm}
\end{figure}
This study is motivated by a financial application. Traders execute trades while observing financial time series images as charts on their desktop screens (Figure~\ref{fig:computer}).
When it comes to financial time series, the data is consumed in its numeric form, but decisions are often augmented by visual representations.

This paper presents a new perspective on numerical time series forecasting by transforming the problem completely into the computer-vision domain.
We capture input data as images and build a network that outputs corresponding subsequent images.
To the best of our knowledge, this is the first study that aims at explicit visual forecasting of time series data as plots. Previous researches leveraged computer vision for time-series data but focused on classifying trade patterns~\cite{cohen2019trading,du2020image}, numeric forecast~\cite{cohen2019effect}, learning weights to combine multiple statistical forecasting methods~\cite{li2020forecasting}, and video prediction for multivariate economic forecasting~\cite{zhen2020video}. 
We follow up on these approaches but focus on an explicit regression-like image prediction task.

This work presents a few advantages. 
Visual time series forecasting is a data-driven non-parametric method, not constrained to a predetermined set of parameters. Thus, the approach is flexible and adaptable to many data forms, as shown by application across various datasets. This bears a stark contrast with classical time series forecasting approaches that are often tailored to the particularity of the data in hand.
The main advantage of this method is that its prediction is independent of other techniques. This is important as it was repeatedly shown that an aggregate of independent techniques outperforms the best-in-class method (e.g.,~\cite{friedman2001elements,goodfellow2016deep,geron2019hands}). 
Secondly, visual predictions result in inherent uncertainty estimates as opposed to pointwise estimates, as they represent distributions over pixels as opposed to explicit value prediction.
In addition, financial time series data are often presented and act upon without having access to the underlying numeric information (e.g., financial trading using the smartphone applications). Thus, it seems viable to examine the value in inferring using visualizations alone.
Lastly, as will be discussed later on, we show that transforming the continuous numeric data to a discrete bounded space using visualization results in robust and stable predictions. We evaluate predictions using multiple metrics. When considering object-detection metrics such as Intersection-over-Union (IoU), visual forecasting outperforms the corresponding numeric baseline. However, when utilizing more traditional time-series evaluation metrics as the symmetric mean absolute percentage error (SMAPE), we find the visual view to perform similarly to its numerical baselines.

\section{Datasets}\label{sec:datasets}
This paper uses four datasets, two synthetic and two real, with varying degrees of periodicity and complexity to examine the utility of forecasting using images.
\begin{figure}[t!]
    \centering
    \includegraphics[width=0.37\textheight]{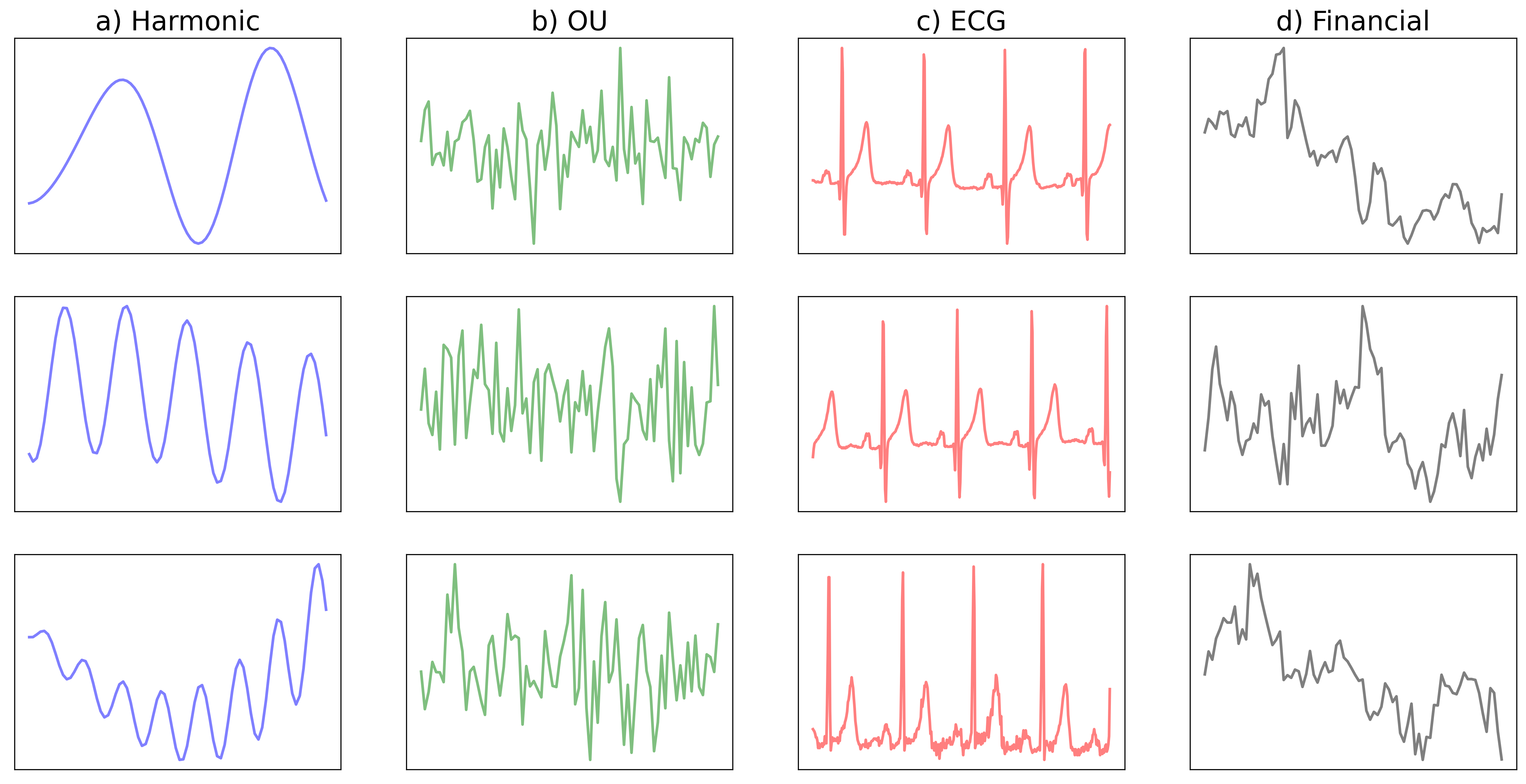}
    \caption{Sampled examples of the four datasets Harmonic, OU, ECG, and Financial.}
    \label{fig:datasets}
\end{figure}
\subsection{Synthetic Data}
We generate two different series for the synthetic datasets: multi-periodic data sampled from harmonic functions and mean-reverting data generated following the Ornstein–Uhlenbeck process.

\subsubsection{\textbf{Harmonic data}}
The first dataset is derived synthetically and is designed to be involved but still with a prominent, repeated signal. We synthesized the time series $s_t$ with a linearly additive two-timescale harmonic generating function,
\begin{equation*}
s_t = (A_1+B_1t)\sin(2\pi t/T_1+\phi_1)
+(A_2+B_2t)\sin(2\pi t/T_2+\phi_2),
\end{equation*}
where the time $t$ varies from $t=1$ to $t=T$, and $T$ denotes the total length of the time series. The multiplicative amplitudes $A_1$ and $A_2$ are randomly sampled from a Gaussian distribution $\mathcal{N}(1, 0.5)$, while the amplitude of the linear trends $B_1$ and $B_2$ are sampled from a uniform distribution $\mathcal{U}(-1/T, 1/T)$. The driving time scales are short ($T_1$) and long ($T_2$) relative to the total length of $T$. Thus, $T_1\sim\mathcal{N}(T/5, T/10)$, while $T_2\sim\mathcal{N}(T, T/2)$. Lastly, the phase shifts $\phi_1$ and $\phi_2$ are sampled from a uniform distribution $\mathcal{U}(0, 2\pi )$. 
We generated and used 42,188 examples as a train set, 4,687 for the validation set, and 15,625 for the test set. Each time series differ concerning the possible combination of tuning parameters.
Panel a) in Figure~\ref{fig:datasets} shows three sampled examples of the harmonic data and it is easy to see that the synthetic time series consist of two time-scales: short oscillations that are composed on a much longer wave trains. 

\subsubsection{\textbf{OU Data}}
We synthesized mean-reverting time series based on Ornstein–Uhlenbeck (OU) process as described in~\cite{byrd2019explaining}. A mean-reverting time series tends to drift towards a fundamental mean value. We chose to synthesize the mean-reverting time series to resemble the characteristics of financial interest rates or volatility. OU's stochastic nature makes it noisy on fine scales but predictable on the larger scale, which is the focus of this study.
Specifically, we generated the OU dataset following the equation adopted from~\cite{byrd2019explaining} with, 
\[s_t \sim \mathcal{N}(\mu + (s_{t-1}-\mu)e^{-\gamma t}, \frac{\sigma^2}{2\gamma}(1-e^{-2\gamma t})),\]
where $\mu$ is the mean value that the time series reverts back to, and $s_0$ starts at $\mu$. We used mean reversion rate $\gamma \sim \mathcal{N}(8e^{-8}, 4e^{-8})$ with units $\text{ns}^{-1}$, and a volatility value $\sigma \sim \mathcal{N}(1e^{-2}, 5e^{-3})$. Overall, we generated the time series by sampling $s_t$ at every minute. 
We generated and used 45,000 examples as a train set, 5,000 for the validation set, and 15,000 for the test set. Similar to the Harmonic data, each time series differ concerning the possible combination of tuning parameters.
Figure~\ref{fig:datasets}(b) shows three samples of the OU data. One can see that the OU data tend to be noisy with uncorrelated ups and downs, but on larger scales, the data is concentrated in the middle of the image as values drift toward the mean due to its reversion constraint. 

\subsection{\textbf{Real Data}}
Along with the synthetically generated data, we use two real-world time series datasets.

\subsubsection{\textbf{ECG data}}
The ECG data is measured information from 17 different people adopted from MIT-BIH Normal Sinus Rhythm Database~\cite{goldberger2000physiobank}. We curated 18 hours of data for each subject after manually examining the data's consistency and validity by analyzing the mean and standard deviation of the time series data for each subject (not shown). 
For each subject, we consider segments of 2.56 seconds (corresponding to 320 data points) sampled randomly from the data. These are then downsampled to 80 data points to be on-par with the other datasets. 13 out of the 17 subjects are used a training data while the other 4 are used as out-of-sample testing data. 
Overall, we sampled 42,188 examples for the training set, while from the test data, we sampled 4,687 as a validation set and 15,625 as a test set.
Panel c) in Figure~\ref{fig:datasets} shows three sampled examples of the ECG data. One can see that the data has prominent spikes about every second, which makes the data predictable. However, there is noticeable noise between spikes that is much harder to predict.

\subsubsection{\textbf{Financial data}}
The last dataset is financial stock data from Yahoo! Finance. The data consists of daily Adjusted Close values of stocks that contributed to the S\&P-500 index since 2000. Each time series segment consists of 80 days and is standardized by subtracting the mean and dividing by the standard deviation for each segment separately.
For the train data, we sampled information randomly from the year 2000 to 2014, while for the test, we sampled information from 2016 to 2019.
Overall, we sampled 38,746 examples as a training set, while from the test data, we sampled 4,306 as a validation set and 15,625 as a test set.

Panel d) in Figure~\ref{fig:datasets} shows three sampled examples of the financial data. Here, one can see that the data is much less predictable than the previous three. 
Although financial data is persistent with sequentially related information, it is hard to spot repeated signals that will make the data predictable. 
Indeed, the prevailing theory of financial markets argues that markets are very efficient, and their future movements are notoriously hard to predict, especially given price information alone (e.g., \cite{pedersen2019efficiently}).

\begin{figure}[t!]
    \centering
    \includegraphics[width=0.36\textwidth]{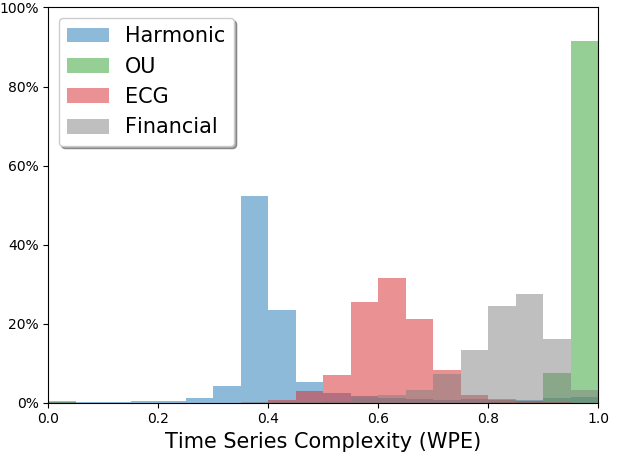}
    \caption{Distribution of dataset's complexity measured using Weighted Permutation Entropy.}
    \label{fig:wpe}
    \vspace{-4mm}
\end{figure}

\subsection{Complexity of Time-Series Data}\label{sec:complexity}
To provide a reference for how the time series across our datasets vary, we measured the complexity of each dataset using a standard measure called Weighted Permutation Entropy~\cite{fadlallah2013weighted} (WPE). 
The larger WPE, the more complex the data is.

As shown in Figure~\ref{fig:wpe}, we can see that the datasets cover a broad range of complexity. As expected, the simplest data is Harmonic with its deterministic periodicity. The ECG data is also periodic but more complex due to its irregularities between the spikes. The Financial data is filled with almost random movement of fine scales, therefore, more complex than both the Harmonic and ECG. The OU data exhibits even more random oscillations and abrupt changes compared to other datasets, thus it is measured as the most complex dataset. However, on the larger scale, the OU data bounces around a hidden mean value with bounded noise, making it possible to predict future value mean and ranges, as we will show later in Section \ref{sec:experiments}.

\section{Problem Statement}\label{sec:prob_statement}
Given a time series signal, our goal is to produce a visual forecast of its future. We approach this problem by first converting the numeric time series into an image (as explained later in Section~\ref{sec:preprocess}) and then producing a corresponding forecast image using deep-learning techniques.

\begin{figure}[t!]
    \centering
    \includegraphics[width=0.46\textwidth]{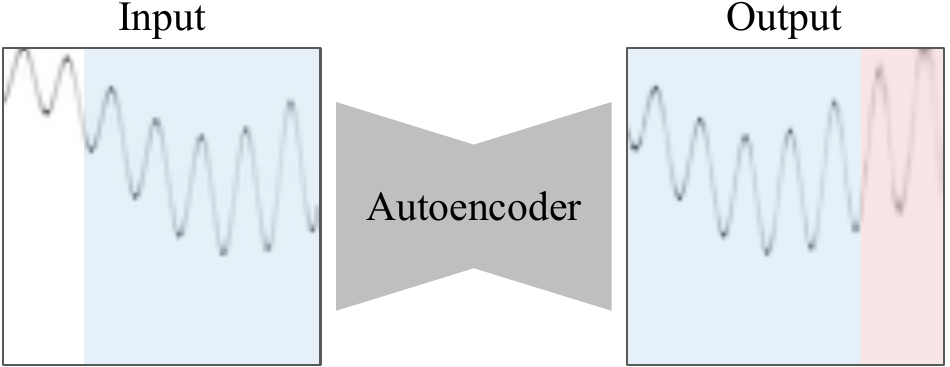}
    \caption{An overview of our problem setup. The blue region shows a 75\% overlap between the input $x$ and output $y$; the forecast region $\hat{y}$ is denoted in red.}
    \label{fig:overview}
    \vspace{-4mm}
\end{figure}

Let $X$ be the set of images of input time series signals, and $Y$ be the set of corresponding forecast output images. The overlap constant $c$ defines the overlap fraction between the input image $x \in X$ and the forecast $y \in Y$, where $c = 1$ implies $x = y, \forall x \in X$, and $c = 0$ implies that $x \cap y = \emptyset, \forall x \in X$, i.e., $x$ and $y$ are distinct. In our experiments, we use $c=0.75$ which means the first 75\% of the forecast image $y$ is simply a reconstruction of the later 75\% of the input image $x$, and the rest 25\% of $y$ corresponds to visual forecasting of the future, as shown in Figure~\ref{fig:overview}. 
We chose $c=0.75$ such that the reconstructed overlap region (first 75\% in $y$) serves as a sanity check on the effectiveness of a forecasting method, and the prediction region (later 25\% in $y$) provides forecasting into the near future. 

\section{Method}
\subsection{Data Preprocessing}\label{sec:preprocess}
Given a 1-d numeric time series $S=[s_0, \cdots, s_T]$ with $s_t \in \mathbb{R}$, we convert $S$ into a 2-d image $x$ by plotting it out, with $t$ being the horizontal axis and $s_t$ being the vertical axis\footnote{We plotted each time series $S$ with bounded intervals. The interval for $x${\em -}axis is $[0-\epsilon, T+\epsilon]$, whereas the interval for $y${\em -}axis is $[\min(s_t)-\epsilon, \max(s_t)+\epsilon]$, where $\epsilon=10^{-6}$.}. 
We standardize each converted image $x$ through following pre-processing steps. First, pixels in $x$ are scaled to $[0,1]$ and negated (i.e., $x = 1 - x/255$) so that the pixels corresponding to the plotted time series signal are bright (values close to 1), whereas the rest of the background pixels become dark (values close to 0). Note that there can be multiple bright (non-zero) pixels in each column due to anti-aliasing while plotting the images.

Upon normalizing each column in $x$ such that the pixel values in each column sum to 1, each column can be perceived as a discrete probability distribution (see Figure~\ref{fig:column-pdf}). Columns represent the independent variable time, while rows capture the dependent variable: pixel intensity. The value of the time series $S$ at time $t$ is now simply the pixel index $r$ (row) at that time (column) with the highest intensity.

Predictions are made over normalized data. To preserve the ability to forecast in physical units, we utilize the span of the input raw data values to transform forecasts to the corresponding physical scales.

\subsection{Image-to-Image Regression}\label{sec:method}

As mentioned in Section \ref{sec:related}, recent work has seen the extensive use of autoencoders in both the time series and computer vision domains. Following these, we extend the use of autoencoders to our image-to-image time series forecasting setting. We use a simplistic convolutional autoencoder to produce a visual forecast image with the continuation of an input time series image, by learning an undercomplete mapping $g \circ f$,
\begin{equation*}
    \hat{y} = g(f(x)),~  \forall x \in X,
\end{equation*}
where the encoder network $f(\cdot)$ learns meaningful patterns and projects the input image $x$ into an embedding vector, and the decoder network $g(\cdot)$ reconstructs the forecast image from the embedding vector. We purposely do not use sequential information or LSTM cells as we wish to examine the benefits of framing the regression problem in an image setting.

\begin{figure}[t!]
    \centering
    \includegraphics[width=0.46\textwidth]{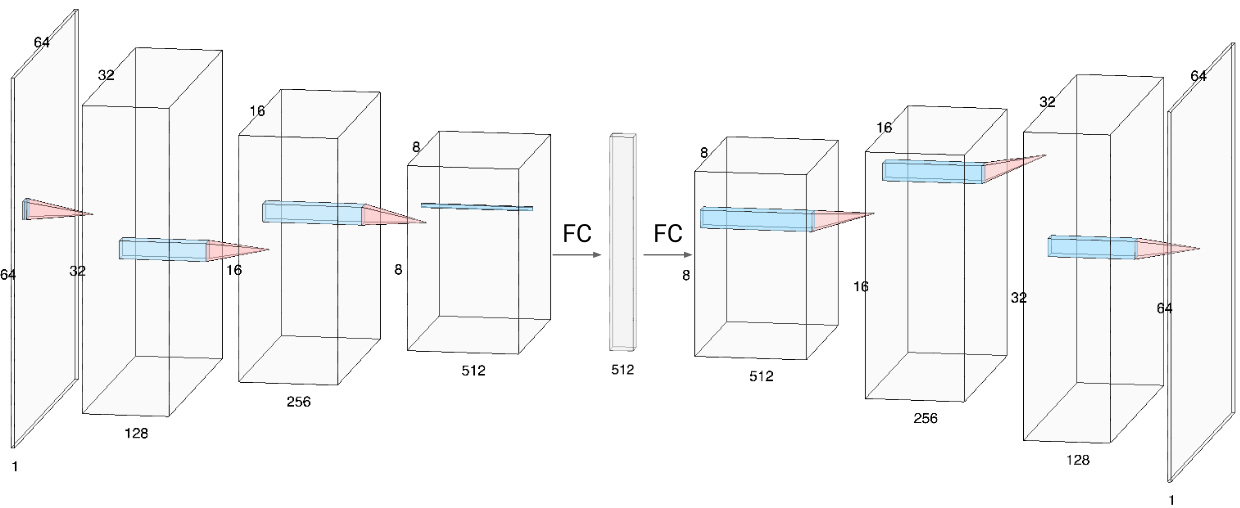}
    \caption{The architecture of undercomplete convolutional autoencoder network used in this study.}
    \label{fig:architecture}
\end{figure}

We call this method \textbf{VisualAE}, the architecture for which is shown in Figure \ref{fig:architecture}. We used 2D convolutional layers with a kernel size of $5\times5$, stride $2$, and padding $2$. All layers are followed by ReLU activation and batch normalization. The encoder network consists of 3 convolutional layers which transform a $80\times80\times1$ input image to $10\times10\times512$, after which we obtain an embedding vector of length $512$ using a fully connected layer. This process is then mirrored for the decoder network, resulting in a forecast image of dimension $80\times80$. We will explain the loss function for training in detail in the next section.

\subsection{Loss Functions}\label{sec:loss_function}
One challenge with the converted time series images is that the majority of the information gets concentrated on fine lines, leaving most of the image blank. This is propagated downstream to the loss function that aims to quantify the dissimilarity of two sparse matrices.

\begin{figure}[t!]
    \centering
    \includegraphics[width=0.45\textwidth]{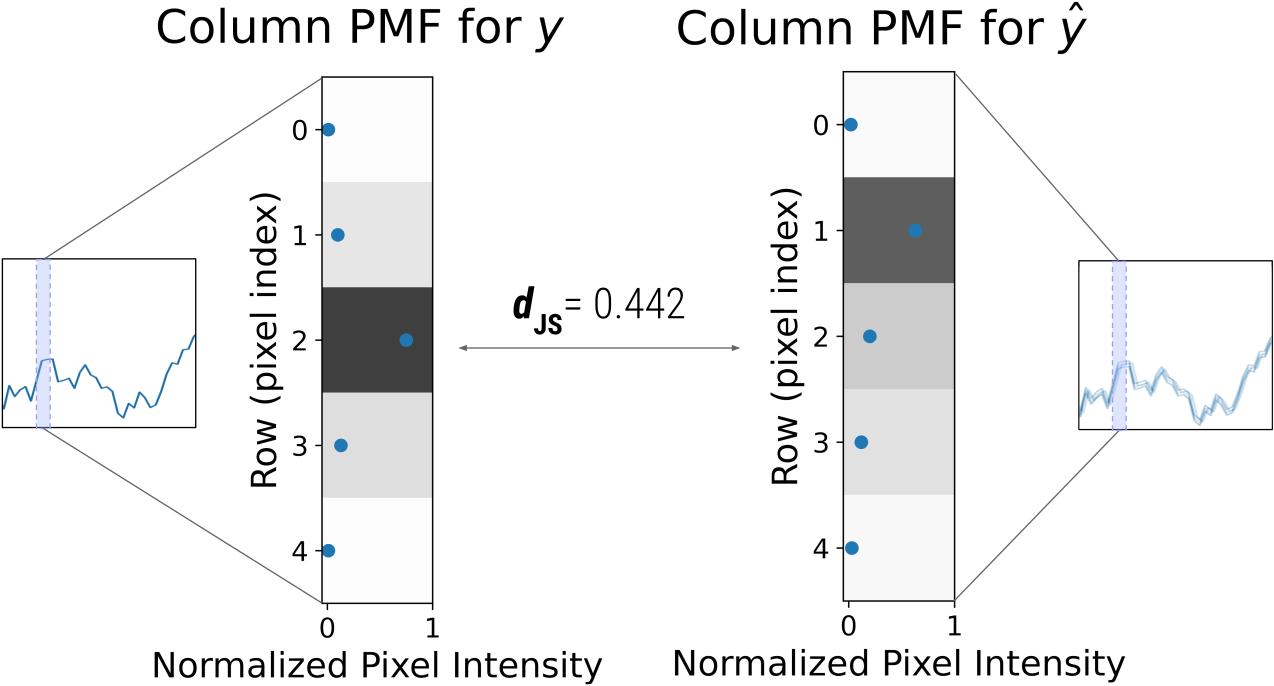}
    \caption{A depiction of comparison of two sample column probability distributions $y=[0.01, 0.1, 0.75, 0.13, 0.01]$ and $\hat{y}=[0.02, 0.63, 0.2, 0.12, 0.03]$.}
    \label{fig:column-pdf}
\end{figure}

We care about the likelihood of pixel intensity in a particular location (row) in each column of the forecast image. This can be achieved by leveraging metrics that compare two probability distributions. We do so in a column-wise manner: the loss $L$ to compare target ground-truth (GT) image $y$ with prediction image $\hat{y}$ is the sum of column-wise distances between the two, 
$$L(y,\hat{y}) = \sum_{i=1}^{w} d(y_{i}, \hat{y_{i}}),$$
where $d$ is any distance measure between two distributions ($y_{i}$ and $\hat{y_{i}}$ in this case), and $w$ is the width of images. This process is depicted in Figure \ref{fig:column-pdf}.


Measures such as the Kullback-Leibler Divergence have been extensively used as loss functions (\cite{goodfellow2016deep}), as they provide a way of computing the distance from an \emph{approximate distribution} $Q$ to a \emph{true distribution} $P$.
In this study, following~\cite{huszar2015not}, we choose $d$ to be the \textbf{Jensen-Shannon Divergence} (JSD), which is a symmetric, more stable version of the Kullback-Leibler Divergence having the property that $D_{JS}(P\|Q) = D_{JS}(Q\|P)$. Here, JSD is computed as
$$D_{JS}(P\|Q) = \frac{1}{2}D_{KL}(P\|M) + \frac{1}{2}D_{KL}(Q\|M) $$
where $M = \frac{1}{2}(P+Q)$. 

\section{Experiments}\label{sec:experiments}

We experimented with four datasets: Harmonic, OU, ECG, and Financial, as they cover a wide range of complexity and predictability in time series data, as discussed in Section \ref{sec:complexity}. In this study we use the PyTorch Lightning framework~\cite{paszke2019pytorch, falcon2019pytorch} for implementation and Nvidia Tesla T4 GPUs in our experiments. We benchmark the proposed method against three baseline methods as we describe below.

As described in Section \ref{sec:prob_statement}, there is a 75\% overlap between input and output (overlap ratio $c=0.75$). Each sample contains 80 datapoints; we aim to forecast the last 20 datapoints (last 25\%) of the output image. This is shown in Figures~\ref{fig:overview} and \ref{fig:qualitative}, where for the predicted image, the first 75\% region (in blue) denotes the reconstructed input, while the last 25\% (in red) region is the visual forecast. All metrics reported are averaged over the unseen red forecast region.

\subsection{Methods}
\begin{table*}[!htb]
\begin{tabular}{|c|c|c|c|c|c|}
\hline
\textbf{} &
  \textbf{Method} &
  \textbf{\begin{tabular}[c]{@{}c@{}}SMAPE\\ $\mu \pm \sigma$\end{tabular}} &
  \textbf{\begin{tabular}[c]{@{}c@{}}MASE\\ $\mu \pm \sigma$\end{tabular}} &
  \textbf{\begin{tabular}[c]{@{}c@{}}IoU\\ $\mu \pm \sigma$\end{tabular}} &
  \textbf{\begin{tabular}[c]{@{}c@{}}JSD\\ $\mu \pm \sigma$\end{tabular}} \\ \hline\hline
\multirow{4}{*}{\textbf{Harmonic}} & RandomWalk        & 1.239 ± 0.440          & 5.106 ± 3.405          & 0.179 ± 0.060          & 0.501 ± 0.043          \\ \cline{2-6} 
                                   & NumAE             & \textbf{0.480 ± 0.297} & \textbf{1.258 ± 1.081} & 0.423 ± 0.107          & 0.334 ± 0.103          \\ \cline{2-6} 
                                   & ARIMA             & 0.580 ± 0.398          & 2.694 ± 3.350          & 0.447 ± 0.238          & 0.343 ± 0.186          \\ \cline{2-6} 
                                   & \textbf{VisualAE} & 0.527 ± 0.303          & 1.518 ± 1.482          & \textbf{0.460 ± 0.088} & \textbf{0.271 ± 0.115} \\ \hline\hline
\multirow{4}{*}{\textbf{OU}}       & RandomWalk        & \textbf{0.018 ± 0.069}          & 1.007 ± 0.385          & 0.257 ± 0.021          & 0.381 ± 0.019          \\ \cline{2-6} 
                                   & NumAE             & \textbf{0.014 ± 0.056} & 471.411 ± 8486.706     & 0.165 ± 0.076          & 0.543 ± 0.052          \\ \cline{2-6} 
                                   & ARIMA             & \textbf{0.014 ± 0.056} & \textbf{0.736 ± 0.133} & 0.141 ± 0.014          & 0.556 ± 0.011          \\ \cline{2-6} 
                                   & \textbf{VisualAE} & \textbf{0.014 ± 0.060} & 0.748 ± 0.119          & \textbf{0.469 ± 0.017} & \textbf{0.257 ± 0.010} \\ \hline\hline
\multirow{4}{*}{\textbf{ECG}}      & RandomWalk        & 1.173 ± 0.463          & 1.551 ± 1.384          & 0.164 ± 0.014          & 0.501 ± 0.021          \\ \cline{2-6} 
                                   & NumAE             & 1.097 ± 0.200          & 0.979 ± 0.280 & 0.278 ± 0.047          & 0.463 ± 0.051          \\ \cline{2-6} 
                                   & ARIMA             & 1.409 ± 0.305          & 1.535 ± 1.688          & 0.160 ± 0.011          & 0.576 ± 0.009          \\ \cline{2-6} 
                                   & \textbf{VisualAE} & \textbf{0.596 ± 0.254} & \textbf{1.658 ± 0.321}          & \textbf{0.485 ± 0.022} & \textbf{0.230 ± 0.041} \\ \hline\hline
\multirow{4}{*}{\textbf{Financial}} &
  RandomWalk &
  \textbf{0.036 ± 0.028} &
  \textbf{3.364 ± 2.217} &
  0.186 ± 0.054 &
  \textbf{0.475 ± 0.050} \\ \cline{2-6} 
                                   & NumAE             & \textbf{0.036 ± 0.028} & \textbf{3.364 ± 2.205} & 0.132 ± 0.069          & 0.598 ± 0.059          \\ \cline{2-6} 
                                   & ARIMA             & \textbf{0.042 ± 0.035}          & 4.034 ± 14.697         & 0.119 ± 0.072          & 0.606 ± 0.063          \\ \cline{2-6} 
                                   & \textbf{VisualAE} & \textbf{0.043 ± 0.028}          & 4.007 ± 2.084          & \textbf{0.212 ± 0.080} & 0.511 ± 0.070          \\ \hline
\end{tabular}
\caption{Summary of various metrics on out-of-sample data with mean ± standard deviation for the forecast region (annotated as red region in Figure~\ref{fig:overview} and Figure~\ref{fig:qualitative}). VisualAE is our proposed method, RandomWalk, NumAE, and ARIMA are the baselines. Lower SMAPE/MASE/JSD error (or higher IoU score) implies better prediction accuracy.}
\label{tab:results}
\vspace{-6mm}
\end{table*}

\subsubsection{\textbf{VisualAE}} This is the proposed method as discussed in Section \ref{sec:method}. We train on images with size $80\times80$. We use a batch size of $128$ and early stopping after 15 consecutive non-improving validation epochs to avoid overfitting during training. We start with a learning rate of $0.1$, which is decayed by a factor of $0.1$ (till $1\mathrm{e}{-4}$) after every 5 non-improving validation epochs.

\subsubsection{\textbf{NumAE} (Numeric AE)}  We also train an autoencoder network to produce numeric forecasts of the original numerical time series signal. The numeric input and output time series are standardized using min-max normalization (with bounds obtained from the input to avoid leakage to future). The autoencoder is trained to predict the output time series by minimizing the Huber loss~\cite{hastie01statisticallearning}.

The architecture, though similar to Figure \ref{fig:architecture}, is shallower (as the dimension of numeric input is much smaller than the images), and uses 1D convolutional layers of kernel size of $5\times5$, stride $2$ and padding $2$. All layers are followed by ReLU activation and batch normalization. The encoder part consists of a series of 2 convolutional layers (of $T/2$ and $T/4$ filters, where $T$ is the length of the signal) and a fully connected layer, which gives us a latent representation of embedding length $T/4$. The decoder is a mirrored encoder.

We use a batch size of $128$, along with a learning rate of $0.01$ which is decayed by a factor of $0.1$ after every 5 consecutive non-improving validation epochs. We also utilize early stopping, as described earlier.

\subsubsection{\textbf{ARIMA}} 
Autoregressive Integrated Moving Average (ARIMA) models are a class of methods that are designed to capture autocorrelations in the data using a combination approach of autoregressive model, moving average model, and differencing (e.g., \cite{wilks2011statistical}). The purpose of each of these three features is to make the model fit the data as well as possible. We used auto arima from pdmaria~\footnote{\url{http://alkaline-ml.com/pmdarima/}} library in our experiments.
    
\subsubsection{\textbf{RandomWalk}} We used the random walk without drift model as a naive numeric forecasting baseline for comparison (e.g., ~\cite{shreve2004stochastic}). Specifically, this model assumes that the first difference of the time series data is not time-dependent, and follows a Gaussian distribution $\mathcal{N}(0, \sigma)$. Given a numeric input time series $\{s_0, \cdots, s_{t-1}, s_t\}$, in order to predict $\{s_{t+1}, \cdots, s_{t+n}\}$, we first estimates $\sigma$ as
\[ \sigma = \sqrt{\mathop{\mathbb{E}}_{i=1}^{t}[(s_i - s_{i-1})^2]}\]
and the prediction at future time $t+k$ follows
\[ s_{t+k} \sim \mathcal{N}(s_t, \sqrt{k}\sigma).\]

This results in a naive numeric forecast that simply extrapolates the last observed value into the future. If we wish to obtain the corresponding image, this forecast is accompanied with a growing uncertainty cone obtained through the equation above.

\subsection{Forecast Accuracy Metrics}\label{sec:metrics}
We use a variety of measures to assess the accuracy of forecast predictions from each method. Some of these metrics are extensively used in the time series forecasting domain, whereas the others we extend from the overarching machine learning field to this task. 

The baseline methods \textbf{ARIMA}, \textbf{NumAE} and \textbf{RandomWalk} produce continuous numeric forecasts, whereas our method \textbf{VisualAE} produces an image. Accordingly, we convert this image back to a numeric forecast which we can use to assess predictions using the metrics described in Section \ref{sec:numeric-metrics}. Similarly, to leverage the image based metrics described in Section \ref{sec:image-metrics}, we transform the numeric predictions of the baseline methods into images using the process described in Section \ref{sec:preprocess}. We discuss the interplay between these metrics across Section~\ref{sec:results}, with further details in Section~\ref{sec:metric-insights}.

\subsubsection{\textbf{Numeric Measures}}\label{sec:numeric-metrics}
\paragraph{\textbf{SMAPE}}
The Symmetric Mean Absolute Percentage Error (or SMAPE) is a widely used measure of forecast accuracy \cite{makridakis2000m3}. It is calculated as:
$$\frac{1}{T} \sum_{i=1}^{T} \frac{|y_{i} - \hat{y_{i}}|}{(|y_{i}| + |\hat{y_{i}}|)/2}$$
where $\hat{y}_i$ is the forecast, $y_i$ the corresponding observed ground-truth, and $t$ is the length of the time series. It ranges from $0.0$ to $2.0$, with lower values indicating better forecasts.

\paragraph{\textbf{MASE}}
The Mean Absolute Scaled Error (or MASE) is another commonly used measure of forecast accuracy \cite{makridakis2000m3}. It is the mean absolute error of a forecast divided by the mean absolute first-order difference of actuals, calculated as
 $$\frac{|e_{j}|}{\frac{1}{T-1}\sum_{i=2}^{T}|y_{i} - y_{i-1}|}$$
where the numerator $|e_{j}|$ is the mean absolute error of the forecast, and the denominator is the mean absolute first-order difference over the ground-truth data period $T$. Errors less than 1 imply that the forecast performs better than the naive one-step method
, with lower values indicating better predictions.

\subsubsection{\textbf{Image based Measures}}\label{sec:image-metrics}
In addition to utilizing tradition forecasting error metrics, we can measure the similarity between the predicted image and the ground-truth image in our setting to evaluate forecast accuracy. We do this through two metrics:

\paragraph{\textbf{JSD}}  Jensen-Shannon Divergence can be used to measure the similarity between image pairs, as described in Section \ref{sec:loss_function}. This ranges from $0.0$ to $1.0$, with lower values indicating a better forecast.

\paragraph{\textbf{IoU}} We extend the Intersection-over-Union (IoU) metric, which is commonly practiced in the object-detection literature~\cite{everingham2010pascal, rezatofighi2019generalized}, to the purpose of measuring forecast accuracy. This ranges from $0.0$ to $1.0$, with higher values indicating better forecasts.

We compute IoU pairwise for each corresponding column in the ground-truth and predicted image. This is done by obtaining the 1D bounding boxes of non-zero pixels for each column and then calculating IoU of corresponding columns from the ground-truth and predicted image. 

\subsection{Results}\label{sec:results}

\begin{figure*}[!ht]
    \centering
    \includegraphics[width=0.75\textwidth]{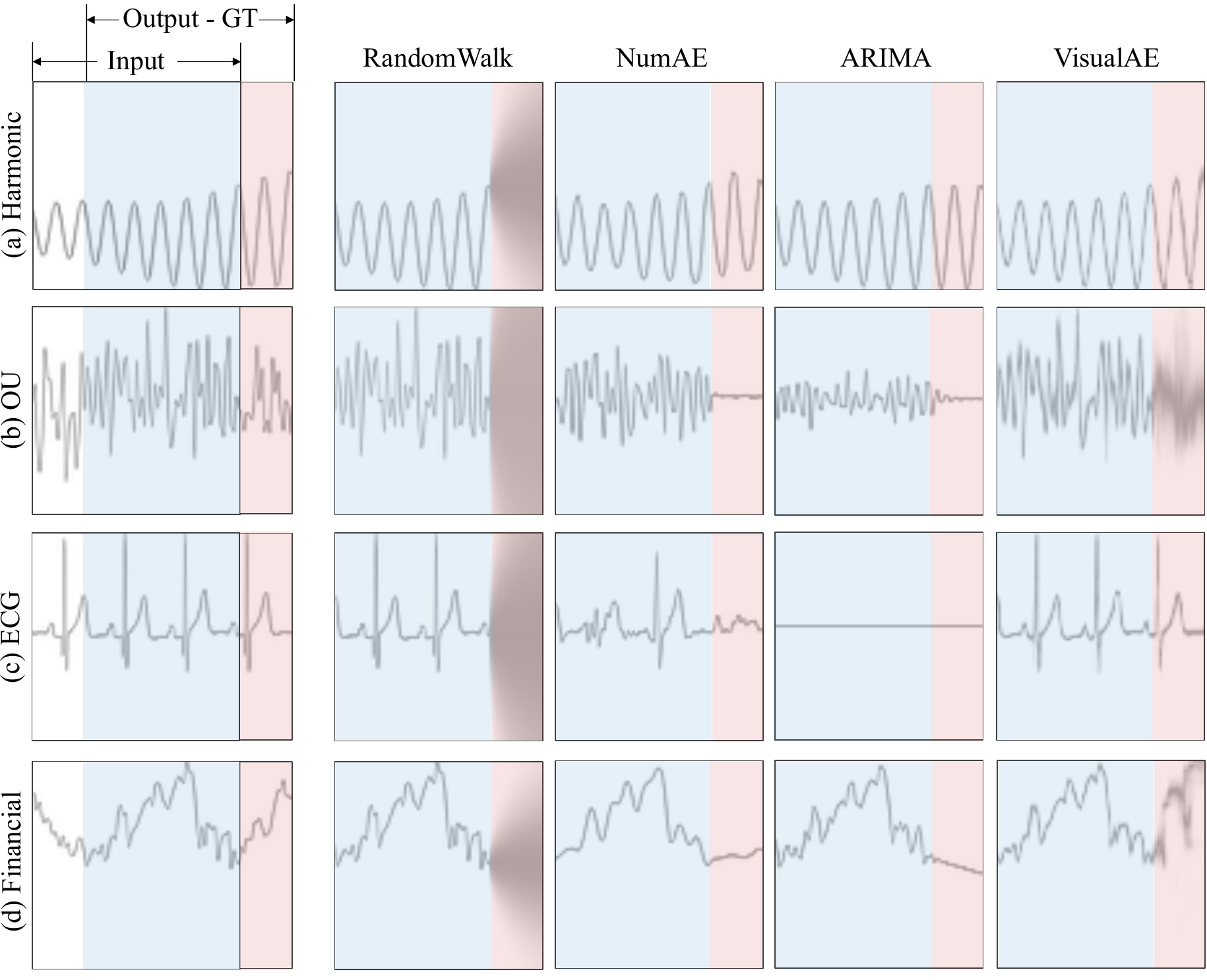}
    \caption{Example out-of-sample forecast predictions using the baseline methods \textbf{RandomWalk}, \textbf{NumAE}, \textbf{ARIMA}, and the proposed method \textbf{VisualAE}. The blue region indicates overlap between the input and output, whereas the red area denotes the future forecast. We show the reconstructed (or fitted) and forecast time series in blue and red region respectively.}
    \label{fig:qualitative}
\end{figure*}

All reported metrics mentioned in Section~\ref{sec:metrics} are over the unseen future prediction region (in red in Figure~\ref{fig:overview}). For both \textbf{VisualAE} and \textbf{NumAE}, we averaged these metrics over five independently trained models with different random weight initializations. We demonstrate that the proposed method \textbf{VisualAE} outperforms baseline methods \textbf{NumAE}, \textbf{RandomWalk}, and \textbf{ARIMA} across all four datasets when evaluated using image-based metrics (such as IoU). However, as we will discuss in this section, traditional numeric metrics are inconsistent with this finding. We demonstrate the value of using a visual approach to time-series forecasting, and how image-based evaluation metrics can help address some of the caveats of traditional numeric metrics. 

We report the mean and standard deviation of various prediction accuracy metrics in Table~\ref{tab:results}. \VisualAE achieves higher IoU scores than all baselines across the four datasets. The same holds true for JSD (with the exception of \RandomWalk scoring better in the Financial dataset). The numeric metrics are often inconsistent – within themselves (SMAPE and MASE) – as well as across the four datasets. According to the numeric metrics, \textbf{VisualAE} is a close second (if not similar) to \textbf{NumAE}, with the exception of the ECG dataset, where \VisualAE performs the best, and the OU dataset, where \ARIMA and \VisualAE perform similarly to \NumAE. We will now discuss the characteristics of benchmarked methods and metrics, along with our overarching findings with the aid of Table~\ref{tab:results} and Figure~\ref{fig:qualitative}.

\begin{figure*}[!htb]
    \centering
    \includegraphics[width=0.85\textwidth]{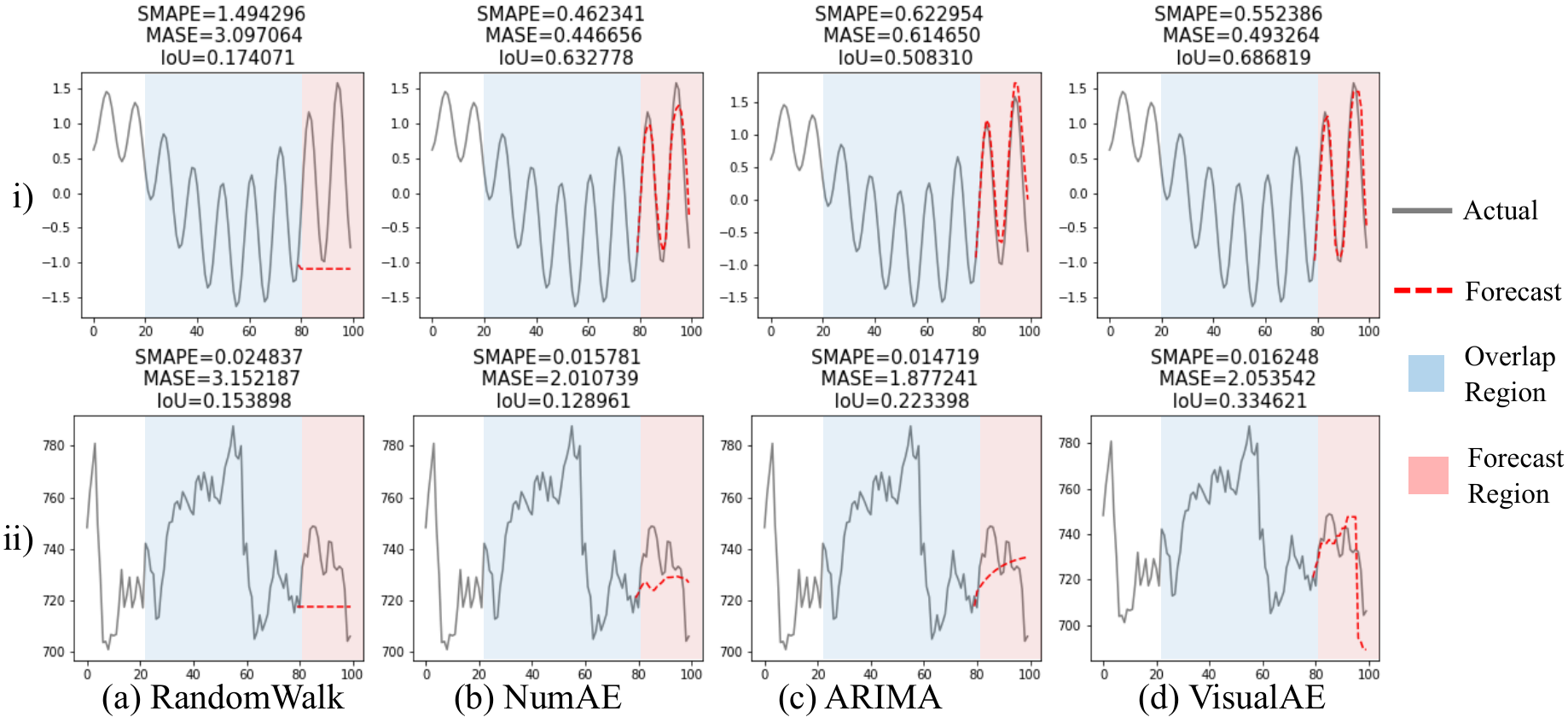}
    \caption{IoU metric better captures visual forecast accuracy compared to traditional numeric metrics SMAPE and MASE.}
    \label{fig:iou-study}
\end{figure*}

\subsubsection{\textbf{Dataset-wise breakdown of approaches}}

\paragraph{\textbf{Harmonic Data:}} The Harmonic dataset are dominated by cyclic patterns (see left column of Figure~\ref{fig:datasets}). As explained in section~\ref{sec:datasets}, each time series in the Harmonic dataset is a mixture (superposition) of two randomly generated individual sinusoids, and each sinusoid exhibits short cyclic patterns along with long term damping or magnifying trends. 
\textbf{ARIMA} performs well but not best because on many occasions, individual sequences don't span the full range of variability needed to tune the model's parameters. This is a barrier for \textbf{ARIMA} as it doesn't have the capability to cross-learn between multiple independent time-series.
The naive forecasting baseline \textbf{RandomWalk} can only learn time-independent stepwise value changes, thus cannot model cyclic patterns. \NumAE and \VisualAE capture these patterns well, as observed in Figure~\ref{fig:qualitative}(a) and Table~\ref{tab:results}, with \NumAE performing slightly better according to numeric metrics, and \VisualAE taking the lead in image-based metrics.


\paragraph{\textbf{OU Data:}} It is hard to predict the exact daily changes in the OU data owing to the random process' nature. However, over a larger scale, the OU data is predictable as a mean-reverting process. Visually, we expect majority of the values to concentrate around the mean value of the time series with some noise. 
\RandomWalk extrapolates the last observed value, whereas \ARIMA extrapolates the in-sample mean value as a steady line for each sample; \NumAE predicts a slightly jagged version of the same. This becomes evident in the abnormally large MASE error, which is sensitive to division by a small term $\epsilon = 1e-4$ when the naive one-step denominator approaches 0. The SMAPE metric appears to be similarly non-informative, as it cannot disambugate the performance of the four methods.
The intricacies of these forecasts are captured in the IoU and JSD metrics, according to which \VisualAE performs the best. This is evident in Figure~\ref{fig:qualitative}(b), where we show that \textbf{VisualAE} concentrates on the hidden mean value, and was also able to partially recover the range of the noise – unlike the other baselines.

\paragraph{\textbf{ECG Data:}} ECG time series are periodical with intermittent spikes, and hence inherently predictable. They have relatively constant frequency and do not posess much time dependent uncertainty. Figure~\ref{fig:qualitative}(c) shows that \textbf{VisualAE} is able to capture these cyclic patterns well, as evidenced by all metrics – image-based and numeric. \VisualAE is able to handle data with sharp and abrupt changes, and better recovers the heart beat spikes as compared to \NumAE. Similar to the Harmonic dataset, \ARIMA is unable to capture the spiky patterns in ECG dataset, and \RandomWalk simply extrapolates the last value.

\paragraph{\textbf{Financial Data:}} Financial time series are the most challenging to forecast amongst the four datasets. According to the prevailing literature (e.g., \cite{pedersen2019efficiently}), financial data is close to random on short scales and shows no apparent periodicity on large scales. Figure~\ref{fig:qualitative}(d) shows that similar to the OU predictions, \textbf{NumAE} and \textbf{ARIMA} predicted the future with a weak linear trend, while \textbf{VisualAE} outperformed with a predicted curve that captures some of the finer details along with the overall nonlinear trend. This is captured by the IoU metric, but if judged according to SMAPE and MASE, \RandomWalk would be the best-performer, tied with \NumAE. This is rather concerning, as solely using numeric metrics would lead us to misleading conclusions, further demonstrating the benefit of using a visual approach in conjunction with traditional numeric methods.

\subsubsection{\textbf{Insights: Numeric vs.~Image based Metrics }}\label{sec:metric-insights}
As discussed in the previous section, numeric metrics are often not consistent with the image-based ones, and sometimes do not agree amongst each other (e.g., see Table~\ref{tab:results}: SMAPE \& MASE values for OU dataset). They are sensitive and often fail to recognize good quality forecasts (e.g., \RandomWalk reportedly performing the best for the Financial dataset). Picking a percentage error such as SMAPE also carries the inability to compare forecast method quality across series (e.g., the low errors in the Financial dataset do not capture that it is in fact the most challenging to predict).

The IoU metric is able to capture this information across the datasets, along with preserving rank-ordering of forecast quality amongst the four methods. Figure~\ref{fig:iou-study}(i) shows an example for the Harmonic dataset, where according to MASE and SMAPE, the \NumAE forecast (column b) is the best performer. This is disputed by IoU, according to which the \VisualAE forecast (column d) is better, and a qualitative visual inspection also suggests the same.
Similarly, in Figure~\ref{fig:iou-study}(ii), we see a hard-to-predict example of the Financial dataset. Just looking at MASE and SMAPE metrics would suggest that both \NumAE and \VisualAE forecasts are of similar quality, whereas a visual inspection shows that \VisualAE captures that long-term trend whereas \NumAE absolutely does not. Once again, the IoU measure captures this difference, reinforcing our belief that a two-pronged approach of utilizing both numeric and visual approaches holds immense value for the field of time series forecasting.

\section{Summary and Conclusion}

To the best of our knowledge, this study is the first to explicitly forecast time series using visual representations of numeric data. 
We show that image-based measures can capture prediction quality more consistently than traditional numeric metrics. The proposed visual forecasting approach, albeit simplistic, performs well across datasets. Our findings show promising results for both periodic time series (including abrupt spikes in ECG) and irregular financial data. We believe that leveraging visual approaches holds immense promise for the field of time series forecasting in the future, especially when used in conjunction with traditional methods.

\subsubsection*{\textbf{Disclaimer:}}
This paper was prepared for information purposes by the Artificial Intelligence Research group of J.~P.~Morgan Chase \& Co.~and its affiliates (“J.~P.~Morgan”), and is not a product of the Research Department of J.~P.~Morgan. J.~P.~Morgan makes no representation and warranty whatsoever and disclaims all liability, for the completeness, accuracy or reliability of the information contained herein.  This document is not intended as investment research or investment advice, or a recommendation, offer or solicitation for the purchase or sale of any security, financial instrument, financial product or service, or to be used in any way for evaluating the merits of participating in any transaction, and shall not constitute a solicitation under any jurisdiction or to any person, if such solicitation under such jurisdiction or to such person would be unlawful.

\bibliographystyle{ACM-Reference-Format}
\bibliography{egbib}

\end{document}